\documentclass[conference]{IEEEtran}

\usepackage[pdftex]{graphicx}
\graphicspath{{../pdf/}{../jpeg/}}
\DeclareGraphicsExtensions{.pdf,.jpeg,.png}

\usepackage[cmex10]{amsmath}
\usepackage{amssymb, amsfonts}

\usepackage[table,xcdraw]{xcolor}   
\usepackage{colortbl}               
\usepackage{multirow}               
\usepackage{booktabs}               

\usepackage{array}
\usepackage{mdwmath}
\usepackage{mdwtab}
\usepackage{eqparbox}

\usepackage{subcaption}
\usepackage{wrapfig}

\usepackage{algorithm}
\usepackage[noend]{algpseudocode}

\usepackage{url}
\usepackage{pifont}
\newcommand{\cmark}{\textcolor{green!60!black}{\ding{51}}}
\newcommand{\xmark}{\textcolor{red}{\ding{55}}}

\begin{document}

\title{\LARGE Accuracy is Not Enough: Poisoning Interpretability in \\ Federated Learning via Color Skew}


 \author{\authorblockN{Farhin Farhad Riya\authorrefmark{1}, Shahinul Hoque\authorrefmark{1}, Jinyuan Stella Sun\authorrefmark{1}, Olivera Kotevska\authorrefmark{2} }
 \authorblockA{\authorrefmark{1} University of Tennessee, Knoxville, USA}
 \authorblockA{\authorrefmark{2} Oak Ridge National Laboratory, USA}}


\maketitle

\begin{abstract}
As machine learning models are increasingly deployed in safety-critical domains, visual explanation techniques have become essential tools for supporting transparency. In this work, we reveal a new class of attacks that compromise model interpretability without affecting accuracy. Specifically, we show that small color perturbations applied by adversarial clients in a federated learning setting can shift a model’s saliency maps away from semantically meaningful regions, while keeping predictions unchanged. The proposed saliency-aware attack framework, called Chromatic Perturbation Module, systematically crafts adversarial examples by altering the color contrast between foreground and background in a way that disrupts explanation fidelity. These perturbations accumulate across training rounds, poisoning the global model’s internal feature attributions in a stealthy and persistent manner. Our findings challenge a common assumption in model auditing that correct predictions imply faithful explanations and demonstrate that interpretability itself can be an attack surface. We evaluate this vulnerability across multiple datasets and show that standard training pipelines are insufficient to detect or mitigate explanation degradation, especially in the FL setting, where stealthy color perturbations are harder to discern. Our attack reduces peak activation overlap in Grad-CAM explanations by up to 35\%, while preserving classification accuracy above 96\% on all evaluated datasets.
\end{abstract}

\IEEEoverridecommandlockouts

\IEEEpeerreviewmaketitle

\section{Introduction}
The growing deployment of machine learning (ML) models in high-stakes domains has underscored the importance of not only model accuracy but also interpretability. Visual explanation techniques, such as Gradient-weighted Class Activation Mapping (Grad-CAM)~\cite{selvaraju2017gradcam}, have become standard tools for interpreting deep learning models, producing heatmaps that highlight regions of an input image most influential in driving predictions. These interpretations support critical functions like auditing, trust calibration, and post-hoc verification.  

Interpretability tools are particularly crucial in Federated Learning (FL)~\cite{kairouz2019advances}, where a global model is trained by aggregating updates from decentralized clients without direct access to their local data. In privacy-sensitive applications such as medical imaging~\cite{sheller2020federated}, smart vehicles~\cite{pokhrel2020federated}, and edge AI~\cite{li2020federated}, post-hoc methods like Grad-CAM are often the only means of verifying whether the model learns meaningful concepts from heterogeneous client distributions. This lack of server visibility makes interpretability indispensable for auditing and trust, but also creates a blind spot where adversarial clients can poison interpretability while preserving accuracy, introducing unique risks for reliability and regulatory compliance. Even when predictions remain correct, their explanations may no longer be semantically justifiable, posing serious risks in safety-critical domains. While prior work has examined adversarial examples and model manipulation \cite{ghorbani2019interpretation, heo2019fooling}, a relatively underexplored threat is whether saliency maps themselves can be systematically distorted during federated training without affecting accuracy. 

In this work, we introduce a new class of stealthy interpretability attacks that exploit this blind spot in FL. The key challenge is to significantly alter saliency maps without changing model predictions. We address this by designing the Chromatic Perturbation Module (CPM), which applies structured, perceptually grounded color transformations to input images~\cite{florence2020see} under a Grad-CAM guided saliency-alignment constraint. Unlike conventional adversarial attacks that alter predictions, CPM preserves classification outcomes while shifting the model’s attention, disrupting interpretability by reducing foreground-background chromatic contrast in salient regions. For each image, perturbation parameters are selected to maintain the original prediction while maximizing dissimilarity, quantified via the Structural Similarity Index Measure (SSIM)~\cite{wang2004image}, between original and perturbed Grad-CAM maps.  

We demonstrate that in FL, adversarial clients can poison interpretability without violating accuracy constraints, exploiting the opacity of the training process. Because servers never see raw client data, color-based perturbations that would be flagged in centralized settings can pass unnoticed, allowing saliency degradation to accumulate over rounds.  

This paper makes the following contributions:
\begin{itemize}
    \item Introduces a new class of interpretability-targeted poisoning attack in FL using structured chromatic perturbations that distort saliency maps without changing predictions.
    \item Demonstrates that such attacks can accumulate over training rounds and persist in the global model.
    \item Provides quantitative and qualitative evidence of saliency degradation while maintaining model accuracy.
    \item Highlights the need for defenses that prioritize explanation fidelity as a core objective in secure FL.
\end{itemize}

\section{Related Work}

Related literature is categorized into three domains to position the work in the broader landscape.

\noindent\textbf{Attacks on Explanation Methods:}  
The first systematic study of attribution robustness was presented by Ghorbani et al.~\cite{ghorbani2019interpretation} that constructed imperceptible perturbations that preserved accuracy while significantly altering saliency maps. Their formulation introduced dissimilarity metrics for capturing saliency misalignment, providing the conceptual foundation for later work. Other approaches include Heo et al.~\cite{heo2019fooling}, who manipulated model weights to degrade interpretability, Chakraborty et al.~\cite{chakraborty2022generalizing}, who evaluated adversarial perturbations under new saliency metrics, and Viering et al.~\cite{viering2019manipulate}, who inserted explicit backdoor patterns. These works demonstrate that interpretability can be manipulated independently of prediction correctness. In contrast, our work departs by focusing on the federated setting, where raw data is not visible to the server, and by introducing structured channel-level color perturbations rather than centralized pixel-level noise. Since such perturbations reduce imperceptibility (adversarial samples are not visually identical to the originals) this attack is particularly suitable for FL, where the server cannot inspect client data and thus cannot easily detect these distortions.

Chen et al.~\cite{chen2019robust} proposed a defense strategy at the pixel level via attribution regularization, while Jyoti et al.~\cite{jyoti2023robustness} surveyed the broader landscape of attribution robustness; however, such pixel-oriented defenses are not suitable for CPM since our perturbations operate at the channel/color level rather than pixel granularity.

\noindent\textbf{Poisoning in Federated Learning:}  
FL has been shown vulnerable to poisoning attacks that target model predictions. Availability attacks reduce global accuracy by injecting corrupted gradients or data~\cite{fang2020local, baruch2019little, bhagoji2019analyzing}, while backdoor attacks implant triggers to cause targeted misclassification without affecting clean performance~\cite{bagdasaryan2020backdoor, xie2020dba, wang2020attack}. Defenses such as KRUM~\cite{blanchard2017machine} and Trimmed Mean~\cite{yin2018byzantine} attempt to filter anomalous updates. However, these methods safeguard predictive performance and do not address the integrity of explanations. Our work complements this literature by demonstrating that even when prediction accuracy is preserved, interpretability can be systematically degraded in FL through adversarial client updates.

\noindent\textbf{Color Perturbations and Interpretability:}  
Vision models are known to be sensitive to chromatic distortions, including hue shifts, brightness changes, and channel rescaling~\cite{engstrom2019exploring}. Some attacks leverage color perturbations to induce misclassifications~\cite{laidlaw2021perceptual}, while others reveal fragility in attribution methods~\cite{ghorbani2019interpretation}. Yet none of these approaches explicitly aim for color perturbations as a stealthy interpretability attack. Our work fills this gap by showing that structured color skew can poison saliency consistency without degrading accuracy, thereby exposing interpretability itself as a new attack surface in decentralized learning, where imperceptibility is not that crucial an adversarial constraint. Our work builds upon and differs from prior literature in three key ways: (i) unlike centralized pixel-level attribution attacks~\cite{ghorbani2019interpretation}, we target the federated setting where imperceptibility is less relevant than stealth against the server; (ii) unlike FL poisoning attacks~\cite{bagdasaryan2020backdoor, fang2020local}, we preserve accuracy while corrupting explanations; and (iii) unlike prior color perturbation studies~\cite{hosseini2018semantic, laidlaw2021perceptual}, we introduce structured channel-level manipulations that accumulate across FL rounds. To the best of our knowledge, this is the first work to systematically study explanation poisoning through realistic color perturbations in FL.

\section{Background}

\noindent\textbf{Federated Learning (FL):} FL enables multiple clients to collaboratively train a global model under a central server without sharing raw data~\cite{mcmahan2017communication}. In each round, selected clients update the model locally and send parameters for aggregation, typically via FedAvg~\cite{mcmahan2017fedavg}. While preserving privacy, FL is vulnerable to poisoning since the server has limited visibility into client data, and heterogeneous (non-IID) distributions further complicate robust training and interpretability.

\noindent\textbf{Visual Explanation Techniques:} Grad-CAM is a widely used post hoc method for CNNs that highlights regions most influential to predictions by weighting feature maps with class gradients~\cite{selvaraju2017gradcam}. It supports debugging~\cite{adebayo2018sanity}, trust~\cite{doshi2017towards}, compliance~\cite{samek2017explainable}, and decision support~\cite{holzinger2019causability}. We emphasize Grad-CAM because its reliance on spatial contrast makes it sensitive to chromatic perturbations introduced by CPM, and its effectiveness depends on consistent, faithful explanations~\cite{adebayo2018sanity, kindermans2019reliability}.

\vspace{-0.5mm}
\section{Threat Model}
\vspace{-0.5mm}
We consider a FL setup consisting of a central server that coordinates $N$ clients, each holding a private, local dataset. The global model is trained over multiple rounds via FedAvg. We assume that a small fraction of clients are malicious and can modify their local data before training.


\subsection{Attacker Model}
\vspace{-0.5mm}
\textbf{Capabilities:} 
Each adversarial client has access to the global model parameters during local training, can compute saliency maps from intermediate gradients and activations, and applies structured color perturbations to its local inputs~\cite{zhao2020idlg}. 

\noindent\textbf{Constraints:} 
The attacker cannot alter ground-truth labels, manipulate model outputs, or access server internals or other clients' data. Perturbations are restricted to the input space and must preserve the predicted class $f(x')=f(x)$. Thus the adversary cannot directly reduce accuracy but only bias interpretability. 

\noindent\textbf{Goals:} 
The objective is to degrade Grad-CAM fidelity by shifting saliency away from meaningful regions while maintaining predictions, allowing distortion to accumulate stealthily across FL rounds. This reflects a low-resource adversary with the same interface as any FL participant, consistent with prior work on stealthy FL poisoning~\cite{bagdasaryan2020backdoor}.

\subsection{Attack Surface and Realism}
\vspace{-0.3mm}
This threat model reflects realistic deployment scenarios, especially in FL applications with heterogeneous hardware (e.g., phones, edge sensors, drones). The decentralized nature of FL means the central server lacks fine-grained visibility into individual client inputs, making this attack more flexible, which can be less imperceptible. The attacker exploits this blind spot to gradually poison the model’s explanation behavior without violating global accuracy metrics.

\section{Attack Design}
\vspace{-0.3mm}
This section details the proposed transformation module, optimization process, and integration into the FL training loop.

\subsection{Chromatic Perturbation Module (CPM)}

The CPM framework is a saliency-aware attack mechanism that generates adversarial inputs $x' \in \mathbb{R}^{H \times W \times 3}$ by applying structured color transformations to the original input $x$, with the goal of perturbing visual interpretability while preserving prediction.

\subsubsection{Foreground-Background Contrast and Saliency Alignment}

We define the \textit{foreground region} $\Omega_f \subseteq \{1, \dots, H\} \times \{1, \dots, W\}$ as the top-$\tau\%$ region in the Grad-CAM saliency map $\text{CAM}(x)$, i.e., \(\displaystyle \Omega_f = \{(i,j) \mid \text{CAM}(x)_{i,j} \geq T_\tau\}, T_\tau=\text{quantile}_{1-\tau}(\text{CAM}(x))
\)and the background region as $\Omega_b = \Omega \setminus \Omega_f$.

Let $\mu_f^c$ and $\mu_b^c$ denote the average pixel intensities in channel $c \in \{R, G, B\}$ over the foreground and background, \(\displaystyle
\mu_f^c = \frac{1}{|\Omega_f|} \sum_{(i,j) \in \Omega_f} x_{i,j}^c, \quad \mu_b^c = \frac{1}{|\Omega_b|} \sum_{(i,j) \in \Omega_b} x_{i,j}^c
\). We define the foreground-background chromatic contrast vector as,
\(\displaystyle
\Delta_\text{fg-bg} = \left[|\mu_f^R - \mu_b^R|,\, |\mu_f^G - \mu_b^G|,\, |\mu_f^B - \mu_b^B|\right]
\).
A high $\|\Delta_\text{fg-bg}\|_2$ indicates strong chromatic separability, which Grad-CAM implicitly leverages when assigning importance.

\subsubsection{Perturbation Operators}

To reduce $\|\Delta_\text{fg-bg}\|_2$ and disrupt saliency localization, we apply differentiable color perturbations to $x$. Here, HSV $(H_x, S_x, V_x)$, corresponding to hue, saturation, and value. 

\begin{itemize}
    \item {Hue Shift:} Applies a global hue rotation $h(\delta)$ in HSV color space where \(\displaystyle
    x' = \text{HSV}^{-1}((H_x + \delta) \mod 1, S_x, V_x)
    \)
    \item {Channel Rescaling:} Modulates channel $c$ via a scale factor $\alpha_c$ where \(\displaystyle
    {x'}_{i,j}^c = \alpha_c \cdot x_{i,j}^c, \quad \alpha_c \in [0.5, 1.5]
    \)
    \item {Contrastive Jitter:} Applies local brightness $\beta$ and contrast $\gamma$ jitter where, \(\displaystyle
    x' = \gamma (x - \mu) + \mu + \beta, \quad \mu = \text{mean}(x)
    \)
\end{itemize}

Each perturbation is parameterized by a vector $\theta = (\delta, \alpha_c, \gamma, \beta)$ and applied as a transformation $\mathcal{T}_\theta(x)$.

\subsubsection{Saliency-Aware Optimization Objective}

For a sample $x$, label $y$, and prediction function $f$, the goal is to find a perturbation $\theta^*$ such that:

\begin{equation}
\begin{aligned}
\theta^* &= \arg\min_\theta \quad \text{SSIM}(\text{CAM}(x), \text{CAM}(\mathcal{T}_\theta(x))) \\
\text{s.t.} \quad & f(\mathcal{T}_\theta(x)) = f(x)
\end{aligned}
\end{equation}

This formulation ensures the model prediction remains unchanged, while the perturbation maximally degrades saliency alignment.

\subsubsection{Implementation Details}

In practice, we discretize the search space of $\theta$ using grid search over a range of hue shifts, channel scalings, and jitter strengths. For each perturbation $\mathcal{T}_\theta(x)$, we compute Grad-CAM, measure its structural similarity to $\text{CAM}(x)$ using SSIM, and select the transformation with the lowest similarity that satisfies the prediction constraint.

\textbf{Remark:} Our attack is thus not stochastic or intensity-agnostic, but rather saliency-driven and foreground-aware. 







\subsection{Federated Strategy}

We assume a synchronous FL setup with $N$ clients, of which a subset $\mathcal{A}$ are adversarial and the rest $\mathcal{B}$ are benign. Let $w_t$ denote the global model at round $t$.

\subsubsection{FedAvg Aggregation}

In each round $t$, a subset $\mathcal{S}_t$ of $K$ clients trains locally for $E$ epochs on private data $\mathcal{D}_i$, producing updates $w_t^i$. The server aggregates them via weighted averaging, \(\displaystyle
w_{t+1} = \sum_{i \in \mathcal{S}_t} \frac{n_i}{\sum_{j \in \mathcal{S}_t} n_j} \, w_t^i,\), where $n_i = |\mathcal{D}_i|$ is the number of local samples.

\subsubsection{Adversarial Update Construction}
Each adversarial client $a \in \mathcal{A} \cap \mathcal{S}_t$ applies the Chromatic Perturbation Module (CPM) to its dataset:
\(\displaystyle\tilde{\mathcal{D}}_a=\{(x', y)\mid x'=\mathcal{T}_{\theta^*(x)}(x), (x, y) \in \mathcal{D}_a\}\), where $\theta^*(x)$ is chosen such that, \(\displaystyle f(x')=f(x),\text{and}\quad\text{SSIM}(\text{CAM}(x), \text{CAM}(x')) \text{is minimized}\) The adversarial local model $w_t^a$ is then trained using $\tilde{\mathcal{D}}_a$, and sent to the server for aggregation, as detailed in Algorithm~\ref{alg:chromatic_attack}.

\subsubsection{Accumulated Saliency Drift}

Let $\mathcal{M}_t$ be the global model after $t$ rounds. Although each adversarial client contributes only a small portion of poisoned data, repeated injection of saliency-targeted perturbations outlined in Algorithm~\ref{alg:chromatic_attack} gradually biases the aggregated model toward distorted representations. Formally, let $x$ be a clean input sample and $\text{CAM}_t(x)$ denote the Grad-CAM map of $x$ under model $\mathcal{M}_t$. The expected structural dissimilarity between rounds increases as \( \displaystyle \mathbb{E}\left[1 - \text{SSIM}(\text{CAM}_0(x), \text{CAM}_t(x))\right] \uparrow \, \text{with} \, t \) while the classification consistency is preserved as \( \displaystyle f_{\mathcal{M}_t}(x) = f_{\mathcal{M}_0}(x), \, \forall x \in \mathcal{D}_{\text{test}}\). We empirically observe that this drift grows approximately linearly with the fraction of adversarial clients and the number of training rounds. Specifically, we find \(\displaystyle
\Delta_t \approx \alpha \cdot r \cdot t
\) where $r$ is the adversarial client ratio, $t$ is the round number, and $\alpha$ is a task-dependent constant. Since each adversarial client trains solely on color-skewed inputs, $r$ also indirectly reflects the proportion of poisoned data in the global update.

\begin{figure}
    \centering
    \includegraphics[width=\columnwidth]{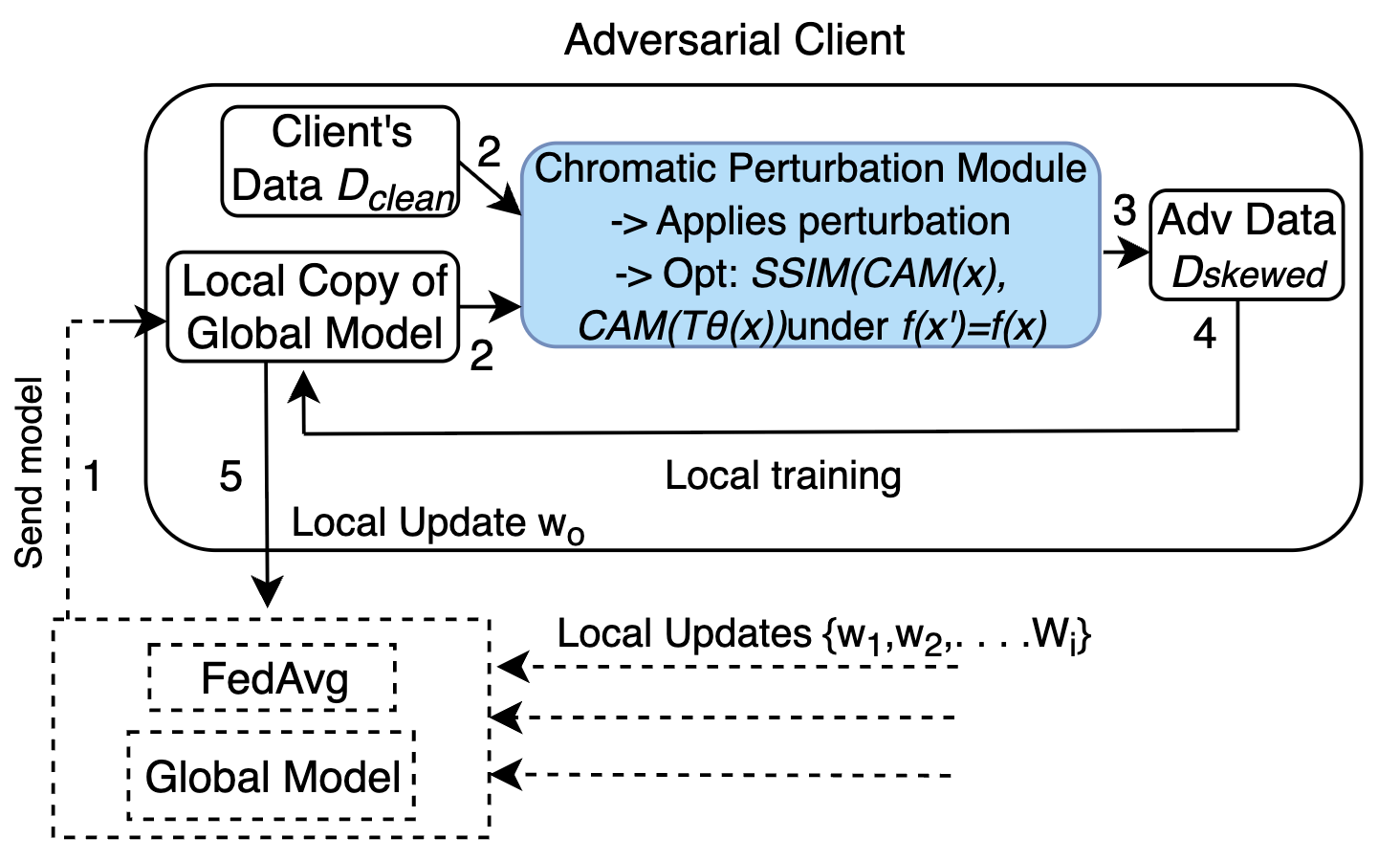}
    \caption{Overview of the attack flow of an adversarial client}
    \label{fig:AttackOverview}
\end{figure}

\begin{algorithm}[htbp]
\caption{Chromatic Saliency Attack in Federated Learning}
\label{alg:chromatic_attack}
\small
\begin{algorithmic}[1]
\State \textbf{Input:} Global model $w_0$, rounds $T$, clients $\mathcal{C} = \mathcal{A} \cup \mathcal{B}$, perturbations $\mathcal{T}$
\For{round $t = 1$ to $T$}
    \State Server selects client subset $\mathcal{S}_t \subset \mathcal{C}$
    \For{client $i \in \mathcal{S}_t$}
        \If{$i \in \mathcal{B}$ (benign)}
            \State Train local model $w_t^i$ on clean data $\mathcal{D}_i$
        \Else
            \State Initialize $\tilde{\mathcal{D}}_i \gets \emptyset$
            \For{sample $(x, y) \in \mathcal{D}_i$}
                \State $\hat{y} \gets f_{w_t}(x)$
                \State $\text{CAM}_{\text{orig}} \gets \text{GradCAM}(x, \hat{y})$
                \For{each perturbation $\theta \in \mathcal{T}$}
                    \State $x_\theta \gets \mathcal{T}_\theta(x)$
                    \If{$f_{w_t}(x_\theta) = \hat{y}$}
                        \State $\text{CAM}_\theta \gets \text{GradCAM}(x_\theta, \hat{y})$
                        \State $s_\theta \gets \text{SSIM}(\text{CAM}_{\text{orig}}, \text{CAM}_\theta)$
                    \EndIf
                \EndFor
                \State $\theta^* \gets \arg\min s_\theta$
                \State $\tilde{\mathcal{D}}_i \gets \tilde{\mathcal{D}}_i \cup (\mathcal{T}_{\theta^*}(x), y)$
            \EndFor
            \State Train local model $w_t^i$ on $\tilde{\mathcal{D}}_i$
        \EndIf
    \EndFor
    \State $w_{t+1} \gets \text{FedAvg}(\{w_t^i\}_{i \in \mathcal{S}_t})$
\EndFor
\State \textbf{Return:} Global model $w_T$
\end{algorithmic}
\end{algorithm}

\section{Evaluation}

To evaluate the efficacy and stealth of our CPM attack, we assess whether Grad-CAM and Grad-CAM++ explanations exhibit semantic drift despite unchanged predictions. The following subsections outline the key evaluation. The extended evaluation is available in the supplementary file.

    
  

\begin{figure}
    \centering \includegraphics[width=0.8\linewidth]{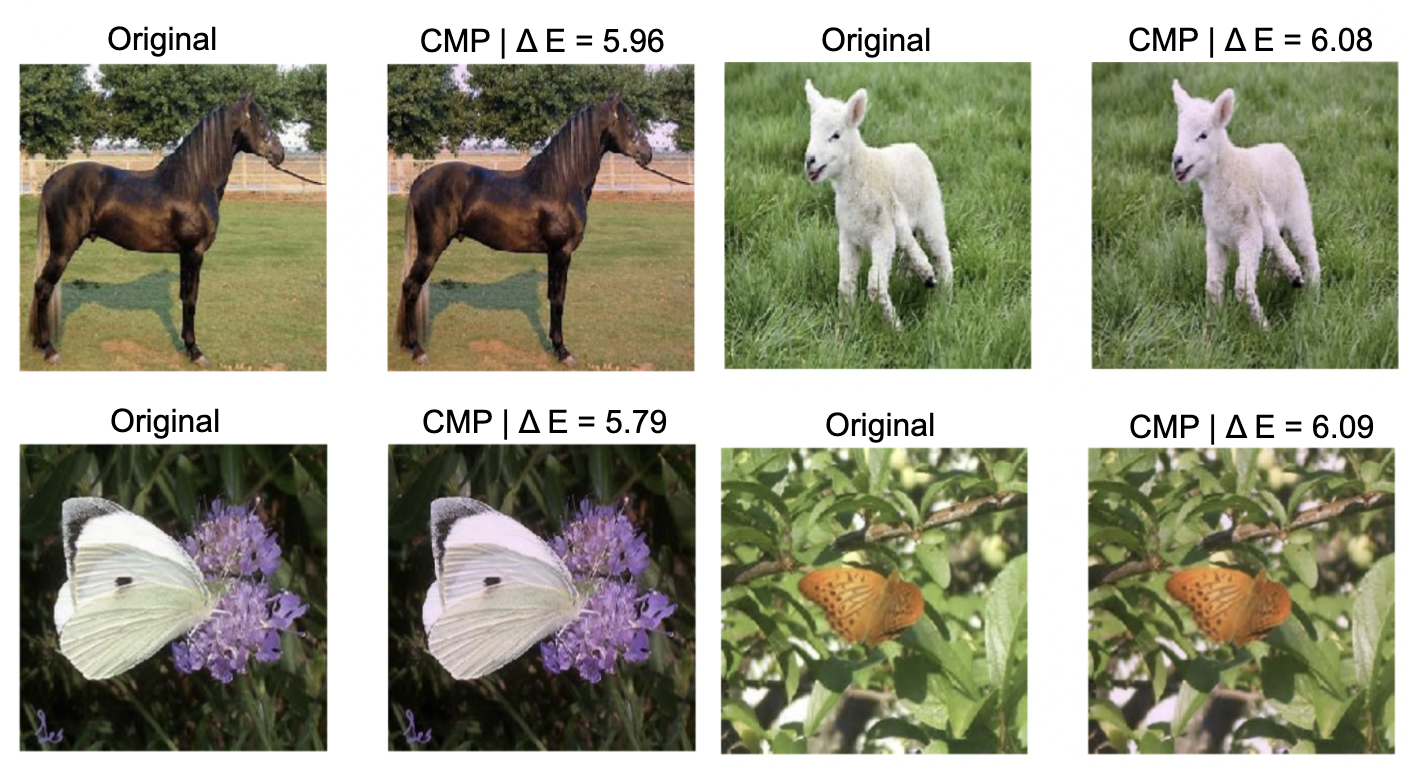}
    \caption{Attack samples generated with CPM. $\Delta E$ values quantify the perceptual color difference (CIEDE2000) between clean and attack samples.}
    \label{fig:attack_samples}
\end{figure}

\begin{figure}
    \centering \includegraphics[width=\columnwidth, height=2.8cm]{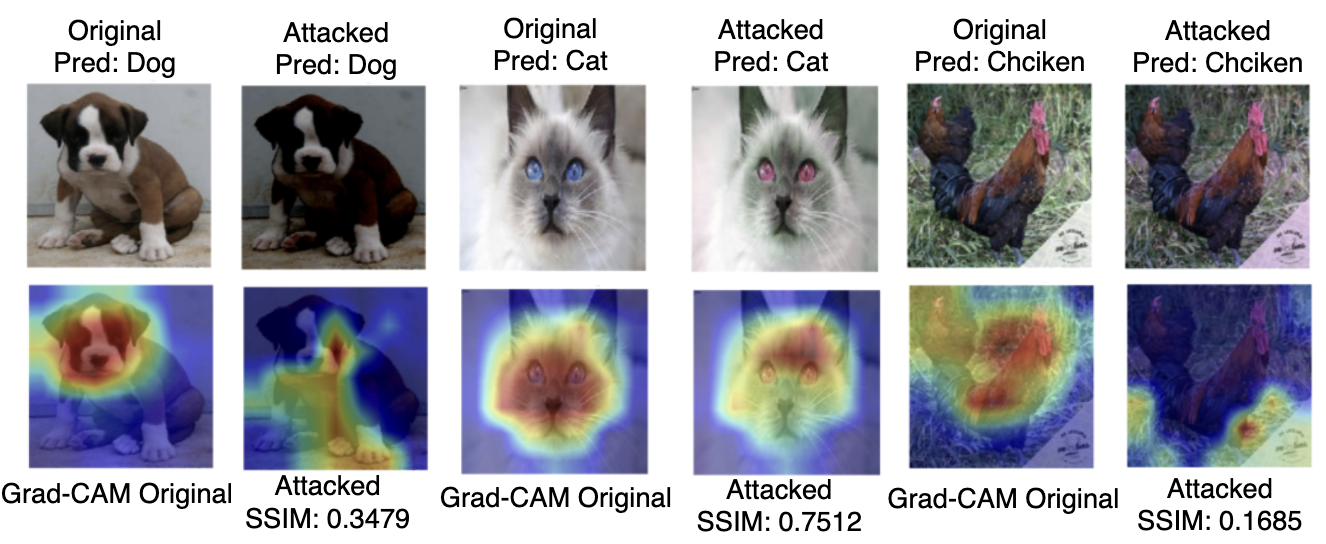}
    \caption{Interpretability of clean model on Attack samples}
    \label{fig:interpratibility_analysis}
\end{figure}

\subsection{Experimental Setup}
All experiments were conducted on Google Colab TPU (T4) using TensorFlow (Python 3.x).
\subsubsection{Model Architecture}
We utilized the \textbf{MobileNet}~\cite{howard2017mobilenets} and \textbf{DenseNet121} \textbf~\cite{huang2017densely} architecture with pre-trained ImageNet weights as the backbone of the baseline model in all settings.

\subsubsection{Dataset}
The evaluation was carried out on the following datasets:
\begin{itemize}
\item \textbf{CIFR-100 and CIFR-10}: CIFR-100 has 100 classes with 50,000 training images and 10,000 test images, whereas CIFR-10 has 10 classes with 50,000 training images and 10,000 for testing. \cite{krizhevsky2009learning}.
\item \textbf{Animal-10}: A dataset of 28,000 images across 10 categories. All images were resized to $224 \times 224$ pixels for training.\cite{corrado2020animals10}.
\item \textbf{Fire}: A binary image classification dataset with $\sim$3000 outdoor images for fire detection 
\cite{phylake2020firedataset}.

\end{itemize}
These datasets were selected to span both general-purpose benchmarks (CIFAR-10, CIFR-100, Animal-10) and safety-critical applications (Fire). In particular, the Fire dataset reflects a crucial Cyber-Physical System (CPS) application, where FL can be leveraged to aggregate edge-device data for robust fire detection. The inclusion of diverse datasets highlights the broader impact and applicability of our attack strategy.


\subsection{Generated Attack Samples}

Figure~\ref{fig:attack_samples} shows examples of CPM-perturbed inputs. Although imperceptibility to humans is not required in FL, we report the average CIEDE2000 color difference $\overline{\Delta E_{00}}$ as a practical reference to prior perceptual robustness studies~\cite{laidlaw2021perceptual} for quantifying perturbation magnitude. Most samples lie in a low-to-moderate (<8), indicating that CPM perturbations are visually plausible rather than extreme. The supplementary material provides full distributions and an ablation relating $\overline{\Delta E_{00}}$ to accuracy and SSIM, along with values for random color perturbations to highlight their uncontrolled nature.

\begin{table}[htbp]
\small 
\setlength{\tabcolsep}{5pt} 
\centering
\begin{tabular}{lccc}
\toprule
\textbf{Dataset} & \textbf{Acc. (\%)} & \textbf{SSIM (↓)} & \textbf{Std.} \\
\midrule
CIFR-100 & 100.0 & 0.501 & 0.0827 \\ 
CIFAR-10      & 100.0 & 0.491 & 0.0832 \\
Animals-10    & 100.0 & 0.482 & 0.0938 \\
Fire          & 100.0 & 0.431 & 0.0781 \\
\bottomrule
\end{tabular}
\caption{Accuracy and SSIM on CPM attack samples for the MobileNet model across all datasets.}
\label{tab:baseline_fidelity}
\end{table}

\subsection{Attack Success Analysis in Baseline Setting}

To assess the generality of CPM, we first evaluate it outside the FL context, simulating a single adversarial client applying CPM on its local model. A clean model trained on unperturbed data is used to generate adversarial samples by perturbing test inputs, ensuring the prediction remains unchanged while maximizing saliency distortion (measured by SSIM). This setup isolates the interpretability impact of CPM without aggregation effects and provides a baseline for later FL comparisons.

\begin{figure}[htbp]
    \centering
    \begin{subfigure}[b]{0.77\columnwidth}
        \includegraphics[width=\columnwidth, height=3cm]{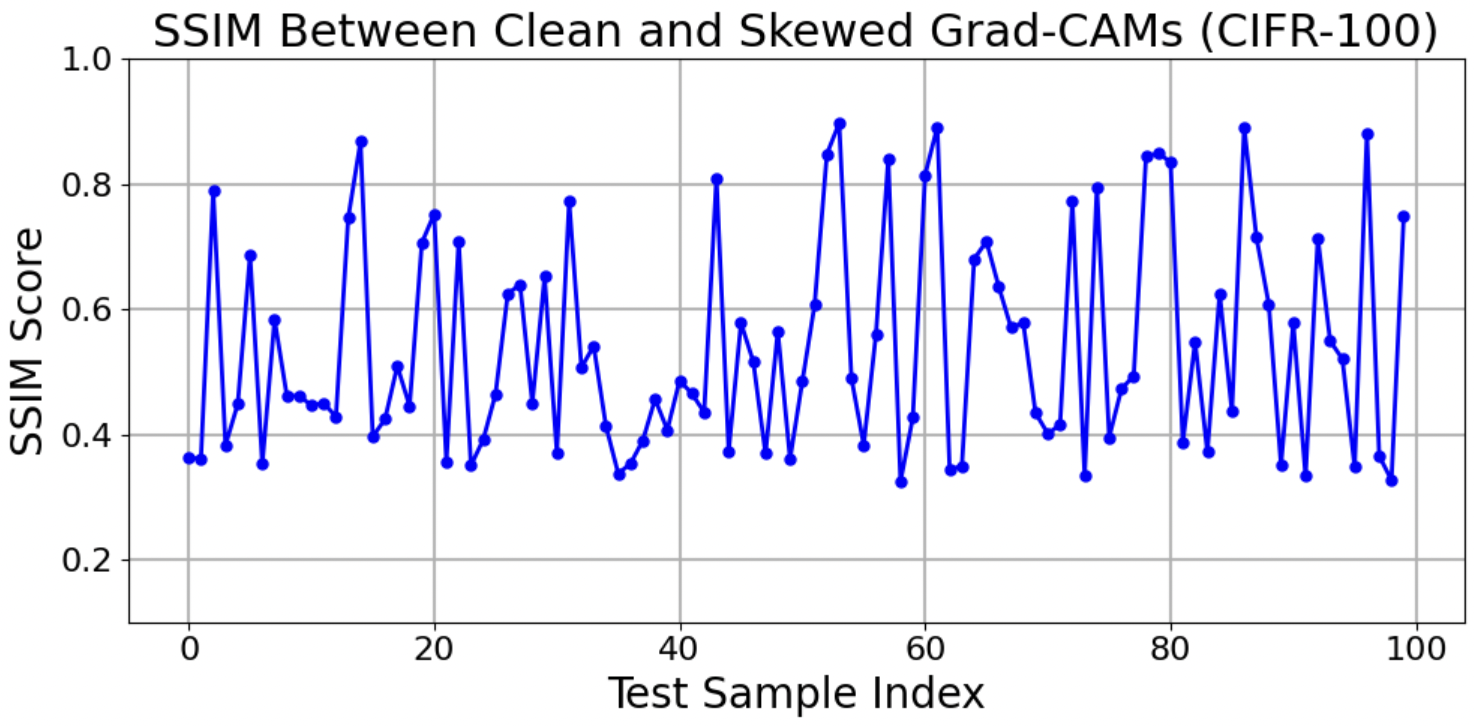}
        \caption{}
        \label{fig:SSIM_100_sample_1}
    \end{subfigure}
    \hfill
    \begin{subfigure}[b]{0.77\columnwidth}
        \includegraphics[width=\columnwidth, height=3cm]{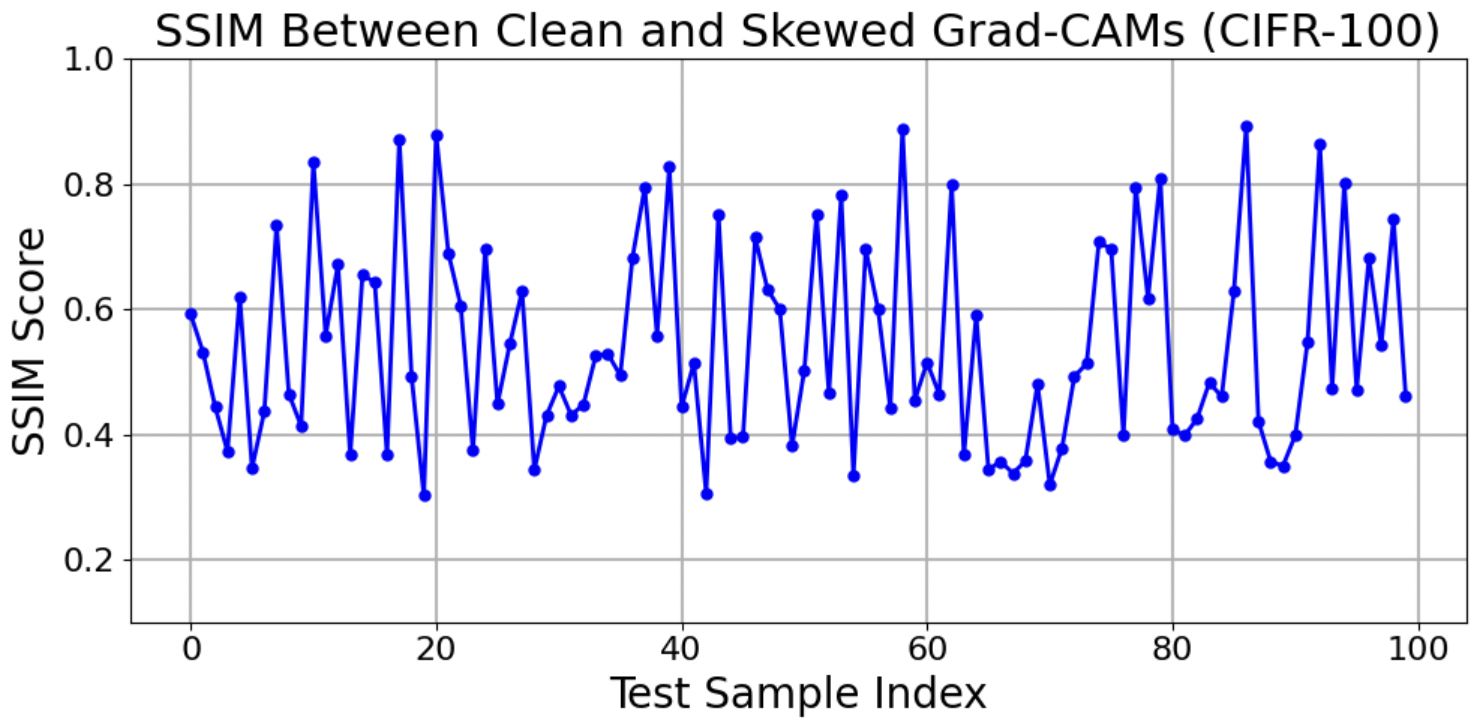}
        \caption{}
        \label{fig:SSIM_100_sample_2}
    \end{subfigure}
    \caption{Comparison of SSIM scores between clean and skewed models using MobileNet (a) and DenseNet121 (b) for CIFR-100 Dataset}
    \label{fig:combined_images2}
\end{figure}

\begin{table}[htbp]
\small
\setlength{\tabcolsep}{2pt}
\centering
\begin{tabular}{p{2.3cm}p{2.8cm}p{2.8cm}}
\toprule
\textbf{Property} & \textbf{Benign Client} & \textbf{Adv. Client} \\
\midrule
Training Data     & Clean (unaltered)      & Perturbed via CPM \\
Label Integrity   & Preserved              & Preserved \\
Model Access      & Local copy        & Local copy   \\
Update Behavior   & Standard training      & Poisoned update \\
Objective         & Model utility (↑)   & Saliency degradation \\
\bottomrule
\end{tabular}

\caption{Comparison of Benign vs. Adversarial Clients}
\label{tab:client_comparison}
\end{table}

\noindent\textbf{Prediction Fidelity:} We assess the fidelity of model predictions under CPM perturbations across the datasets. Adversarial counterparts of all the test samples were generated using CPM. As shown in Table \ref{tab:baseline_fidelity}, all perturbed inputs per dataset were classified identically to their clean counterparts, confirming that CPM preserves decision boundaries while applying perceptually benign distortions.

\noindent\textbf{Interpretability Analysis:} While predictions remain intact, the saliency maps generated for perturbed inputs differ significantly from those of the original clean samples. Lower SSIM scores indicate greater dissimilarity in explanation. As illustrated in Figure~\ref{fig:interpratibility_analysis}, CPM causes visible and semantically significant saliency drift, even though the classification outcome is preserved. These results demonstrate that interpretability can be compromised at the client level before federated aggregation. Figure~\ref{fig:SSIM_100_sample_1} and Figure~\ref{fig:SSIM_100_sample_2} report SSIM scores on 100 test samples (for clear visualization) using MobileNet and DenseNet121, respectively. In both cases, over 45\% of samples exhibit SSIM below 0.5, with some as low as 0.3, indicating substantial interpretability degradation even without prediction change.

\begin{table*}[htbp]
\centering
\renewcommand{\arraystretch}{1.0}
\setlength{\tabcolsep}{8pt}
\resizebox{\textwidth}{!}
{%
\begin{tabular}{|c|c|c|c|c|c|c|c|}
    \hline
    \textbf{Dataset} & \textbf{Rd.} & \textbf{Adv.(\%)} 
    & \textbf{Pre.Fd.(\%)} 
    & \textbf{SSIM(GC/GC++)} 
    & \textbf{Std(GC/GC++)} 
    & \textbf{Peak(\%)} 
    & \textbf{L1} \\
    \hline \hline

    \multirow{8}{*}{CIFAR-10} 
    & \multirow{4}{*}{20} 
        & 0\% \cellcolor{gray!20}& 94.2 \cellcolor{gray!20}& 1.000/1.000 \cellcolor{gray!20}& 0.000/0.000 \cellcolor{gray!20}& 100.0 \cellcolor{gray!20}& 0.00 \cellcolor{gray!20}\\
        &      & 10\% & 93.8 & 0.745/0.712 & 0.062/0.067 & 72.4 & 0.25 \\
        &      & 30\% & 94.1 & 0.611/0.580 & 0.074/0.076 & 58.1 & 0.41 \\
        &      & 50\% & 93.9 & 0.552/0.519 & 0.082/0.085 & 50.7 & 0.50 \\
    \cline{2-8}
    & \multirow{4}{*}{50} 
        & 0\%  \cellcolor{gray!20}& 98.2 \cellcolor{gray!20}& 1.000/1.000 \cellcolor{gray!20}& 0.000/0.000 \cellcolor{gray!20}& 100.0 \cellcolor{gray!20}& 0.00 \cellcolor{gray!20}\\
        &      & 10\% & 98.7 & 0.630/0.595 & 0.063/0.065 & 66.3 & 0.31 \\
        &      & 30\% & 98.1 & 0.389/0.351 & 0.078/0.081 & 41.2 & 0.61 \\
        &      & 50\% & 98.6 & 0.317/0.284 & 0.084/0.088 & 35.5 & 0.73 \\
    \hline

    \multirow{8}{*}{Animals-10} 
    & \multirow{4}{*}{20} 
        & 0\%  \cellcolor{gray!20}& 97.8 \cellcolor{gray!20}& 1.000/1.000 \cellcolor{gray!20}& 0.000/0.000 \cellcolor{gray!20}& 100.0 \cellcolor{gray!20}& 0.00 \cellcolor{gray!20}\\
        &      & 10\% & 95.3 & 0.762/0.728 & 0.049/0.053 & 73.5 & 0.22 \\
        &      & 30\% & 96.6 & 0.628/0.598 & 0.065/0.068 & 59.7 & 0.37 \\
        &      & 50\% & 96.2 & 0.579/0.543 & 0.077/0.079 & 52.3 & 0.46 \\
    \cline{2-8}
    & \multirow{4}{*}{50} 
        & 0\%  \cellcolor{gray!20}& 98.8 \cellcolor{gray!20}& 1.000/1.000 \cellcolor{gray!20}& 0.000/0.000 \cellcolor{gray!20}& 100.0 \cellcolor{gray!20}& 0.00 \cellcolor{gray!20}\\
        &      & 10\% & 97.7 & 0.633/0.601 & 0.051/0.055 & 69.8 & 0.28 \\
        &      & 30\% & 97.5 & 0.423/0.387 & 0.069/0.072 & 47.5 & 0.54 \\
        &      & 50\% & 97.8 & 0.350/0.315 & 0.080/0.083 & 39.6 & 0.65 \\
    \hline

    \multirow{8}{*}{Fire} 
    & \multirow{4}{*}{20} & 0\%  \cellcolor{gray!20}& 96.4 \cellcolor{gray!20}& 1.000/1.000 \cellcolor{gray!20}& 0.000/0.000 \cellcolor{gray!20}& 100.0 \cellcolor{gray!20}& 0.00 \cellcolor{gray!20}\\
        &      & 10\% & 96.0 & 0.812/0.785 & 0.054/0.059 & 78.0 & 0.19 \\
        &      & 30\% & 95.2 & 0.693/0.658 & 0.068/0.072 & 64.1 & 0.35 \\
        &      & 50\% & 95.8 & 0.610/0.579 & 0.072/0.076 & 56.2 & 0.44 \\
    \cline{2-8}
    & \multirow{4}{*}{50} 
        & 0\% \cellcolor{gray!20}  & 98.4 \cellcolor{gray!20} & 1.000/1.000 \cellcolor{gray!20}& 0.000/0.000 \cellcolor{gray!20}& 100.0 \cellcolor{gray!20}& 0.00 \cellcolor{gray!20}\\
        &      & 10\% & 94.9 & 0.651/0.620 & 0.057/0.061 & 70.9 & 0.26 \\
        &      & 30\% & 95.1 & 0.429/0.398 & 0.069/0.073 & 49.8 & 0.48 \\
        &      & 50\% & 95.6 & 0.351/0.319 & 0.074/0.078 & 41.3 & 0.59 \\
    \hline

    \multirow{8}{*}{CIFR-100} 
    & \multirow{4}{*}{20} 
         & 0\% \cellcolor{gray!20} & 92.5\cellcolor{gray!20}  & 1.000/1.000 \cellcolor{gray!20} & 0.000/0.000 \cellcolor{gray!20} & 100.0 \cellcolor{gray!20} & 0.00 \cellcolor{gray!20} \\
        &      & 10\% & 92.8 & 0.693/0.662 & 0.071/0.076 & 68.9 & 0.33 \\
        &      & 30\% & 92.1 & 0.541/0.509 & 0.083/0.087 & 52.7 & 0.52 \\
        &      & 50\% & 92.9 & 0.463/0.429 & 0.089/0.094 & 44.8 & 0.63 \\
    \cline{2-8}
    & \multirow{4}{*}{50} 
         & 0\% \cellcolor{gray!20} & 95.6 \cellcolor{gray!20} & 1.000/1.000 \cellcolor{gray!20} & 0.000/0.000 \cellcolor{gray!20} & 100.0 \cellcolor{gray!20} & 0.00 \cellcolor{gray!20} \\
        &      & 10\% & 95.4 & 0.582/0.547 & 0.078/0.081 & 61.2 & 0.41 \\
        &      & 30\% & 95.8 & 0.403/0.369 & 0.084/0.088 & 41.4 & 0.66 \\
        &      & 50\% & 95.1 & 0.408/0.371 & 0.092/0.097 & 40.7 & 0.78 \\
    \hline
\end{tabular}}
\caption{Federated evaluation of the Chromatic Perturbation Module (CPM) across datasets. Top rows under each dataset indicate clean models (0\% adversarial clients). Prediction Fidelity (Pre.Fd.) reports the percentage of prediction consistency compared to Vanilla FL. SSIM is computed between Grad-CAM and Grad-CAM++ heatmaps of clean vs. poisoned models. Peak Overlap (\%) quantifies focus alignment; L1 Distance measures saliency shift.}
\label{tab:fl_attack_full_metrics}
\end{table*}

\begin{figure}[htbp]
    \centering
    \begin{subfigure}[b]{\linewidth}
        \includegraphics[width=\columnwidth]{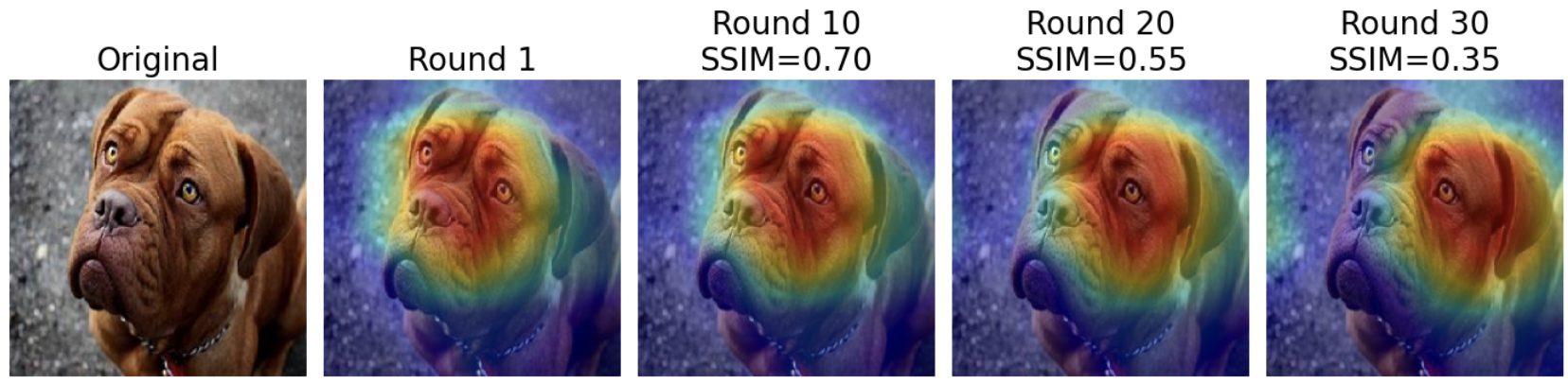}
        \label{fig:accumulation1}
    \end{subfigure}
    \begin{subfigure}[b]{\linewidth}
        \includegraphics[width=\columnwidth]{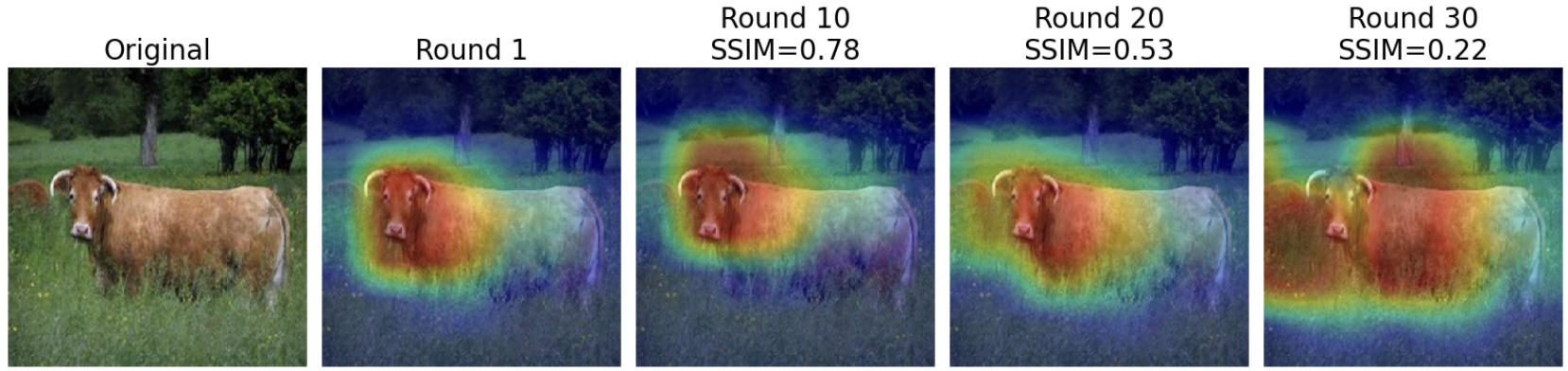}
        \label{fig:accumulation2}
    \end{subfigure}
    \caption{Visualization of the accumulated distortion of Grad-CAM explanations under CPM across FL rounds}
    \label{fig:accumulation_main}
\end{figure}

\begin{figure*}
    \centering
    \includegraphics[width=\linewidth, height=5cm]{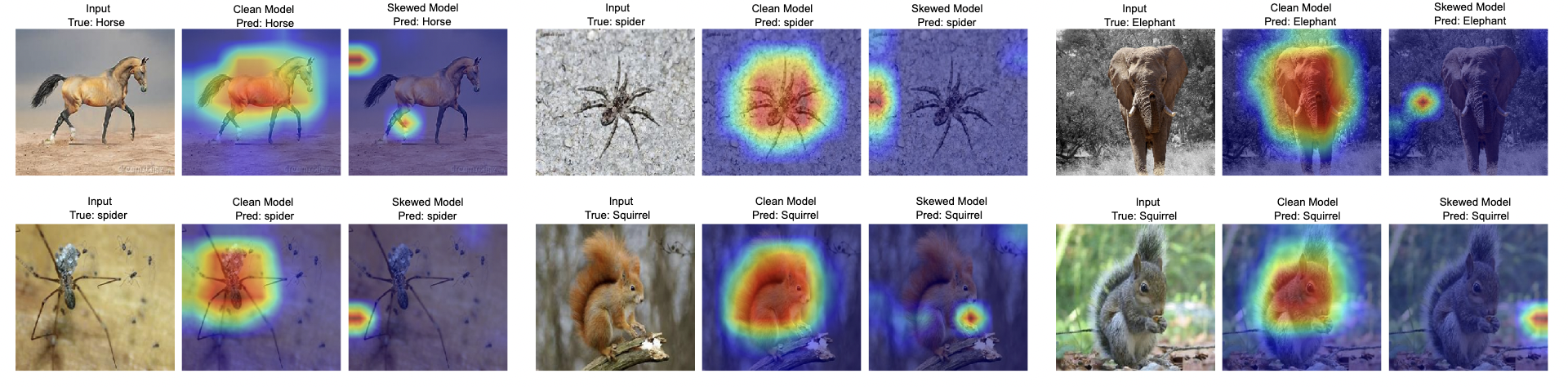}
    \caption{Interpretability analysis of Clean vs Skewed model (Adversarial Client Ration over ~80\%) on clean test samples where the SSIM score drops as low as 2\%}
    \label{fig:combined_images}
\end{figure*}

\begin{figure}[htbp]
    \centering
    \includegraphics[width=0.75\columnwidth]{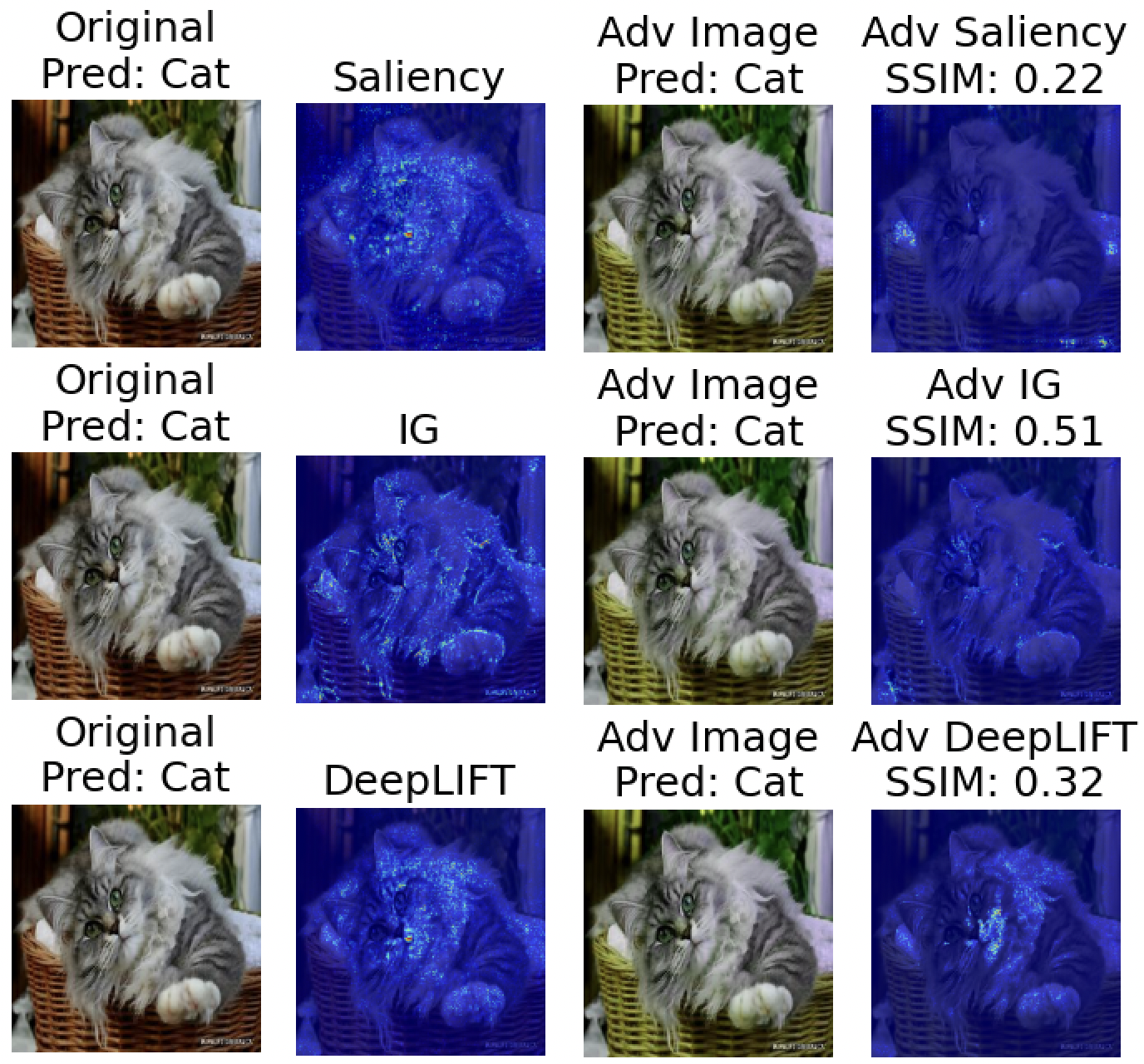}
    \caption{Effect of CPM attack on non-CAM attribution methods}
    \label{fig:3methodCMP}
\end{figure}


\subsection{Attack Success Analysis in FL Setting}

In this section, we evaluated the Grad-CAM drift over multiple FL rounds, starting from a clean model and injecting poisoned updates from different fractions of adversarial clients. Table~\ref{tab:client_comparison} summarizes the key behavioral differences between benign and adversarial clients in our attack setting.

\noindent\textbf{Vanilla FL:}To contextualize our results, we first report the baseline global model accuracy without adversarial clients or CPM.  
Across 20 and 50 communication rounds, the vanilla FL models achieve $64.7\% \to 70.2\%$ accuracy on CIFAR-10, $85.5\% \to 90.3\%$ on Animals-10, $90.2\% \to 92.1\%$ on Fire, and $61.3\% \to 61.8\%$ on CIFAR-100.  
These numbers represent the true predictive capability of the global model in the absence of any attack.  

\noindent\textbf{Prediction Fidelity:} Prediction Fidelity quantifies the stability of the model's accuracy under CPM perturbations. For each dataset, we evaluate the baseline global model trained with vanilla FL and then compute the percentage of test predictions that remain identical between clean inputs and their corresponding CPM-perturbed versions. Although all adversarial samples generated by CPM individually preserve accuracy (Table \ref{tab:baseline_fidelity}), fidelity values are slightly below 100\% because federated training involves random client selection, where data diversity across rounds introduces accuracy fluctuations. 
As reported in Table~\ref{tab:fl_attack_full_metrics}, fidelity consistently exceeds 95\% across all datasets and rounds, confirming that CPM preserves decision boundaries while substantially degrading interpretability.

\noindent\textbf{Saliency Drift Analysis:}
Despite unchanged predictions, saliency fidelity measured via SSIM between Grad-CAM and Grad-CAM++ ~\cite{chattopadhay2018gradcam++} declines sharply as the proportion of adversarial clients increases. In CIFAR-10, SSIM (GC++) drops from 1.000 to 0.284 at 50\% adversaries by Round 50. Peak Overlap (top-$k$ overlap pixels) falls from 100\% to 35\%, indicating a major shift in model focus. L1 distances further confirm this saliency deviation. To illustrate, Figure~\ref{fig:combined_images} compares Grad-CAM heatmaps from clean and skewed models (with over 80\% adversarial clients). Even on clean inputs, the skewed model's explanations diverge significantly, with SSIM as low as 0.02 and averages below 0.10, demonstrating that the attack causes both statistically and visually disruptive shifts in explanation, especially concerning in safety-critical settings.

While there is no universal threshold for when a model becomes untrustworthy due to explanation degradation, our results suggest that significant interpretability loss occurs when SSIM between clean and poisoned Grad-CAM heatmaps drops below 0.4 and peak overlap falls under $\sim$ 50\%.

\noindent\textbf{Accumulation Across FL Rounds:}
The attack’s impact intensifies over time, as seen in the progression from FL Round 20 to Round 50. Saliency metrics worsen with each round due to the cumulative nature of poisoned updates. For example, in the Fire dataset, SSIM (GC++) drops from 0.785 to 0.620 between Round 20 and 50 at just 10\% adversarial clients, while L1 distance increases from 0.19 to 0.26. This illustrates how even moderate adversarial presence can compound over time, degrading explanation quality without noticeable changes in accuracy. These findings demonstrate that current FL aggregation mechanisms like FedAvg are susceptible to long-term interpretability attacks unless explanation fidelity is explicitly monitored.
To visualize the distortion of Grad-CAM heatmaps under CPM, Figure \ref{fig:accumulation_main} contains two representative samples across federated rounds. For both cases, in the early rounds (Round 10), the explanation remains close to the original, with high SSIM values ($\sim0.7$). As training progresses, the perturbations accumulate in the global model updates and steadily displace the saliency focus away from the relevant regions. By Round 30, SSIM has dropped as low as 0.22.

\subsection{Random Color Skew}
\vspace{-0.5 em}
CPM optimizes perturbations under the constraint \(f(x')=f(x)\), where \(x'=\mathcal{T}_\theta(x)\), ensuring that predictions remain unchanged. As a baseline, we implement random color skewing, where perturbations are sampled without optimization across hue, saturation, and channel scales. Formally, each pixel channel is perturbed as  
\(\displaystyle
x' = x + \delta, \ \delta \sim \mathcal{N}(0,\sigma^2) \ \text{with projection into HSV/RGB space,}\)  where \(\delta\) is applied as a random hue shift \( \mathcal{U}[-30^\circ,30^\circ])\), saturation scaling \(\mathcal{U}[0.5,1.5])\), channel rescaling \(\mathcal{U}[0.8,1.2])\), either individually or in random combinations. Unlike CPM, which enforces prediction consistency, uncontrolled skews frequently alter predictions and yield unstable interpretability.  

\noindent\textbf{Decision Boundaries:} For a class region \(\mathcal{R}_y = \{x \mid f(x)=y\}\), random color skews can push inputs outside \(\mathcal{R}_y\), resulting in misclassifications. In contrast, CPM explicitly searches within \(\mathcal{R}_y\), preserving labels while maximizing shifts in heatmaps. This makes CPM a targeted interpretability attack, whereas random color skew serves only as an uncontrolled baseline.  

\begin{figure}[htbp]
    \centering
    \begin{subfigure}{0.54\columnwidth}
        \centering
        \includegraphics[width=\columnwidth]{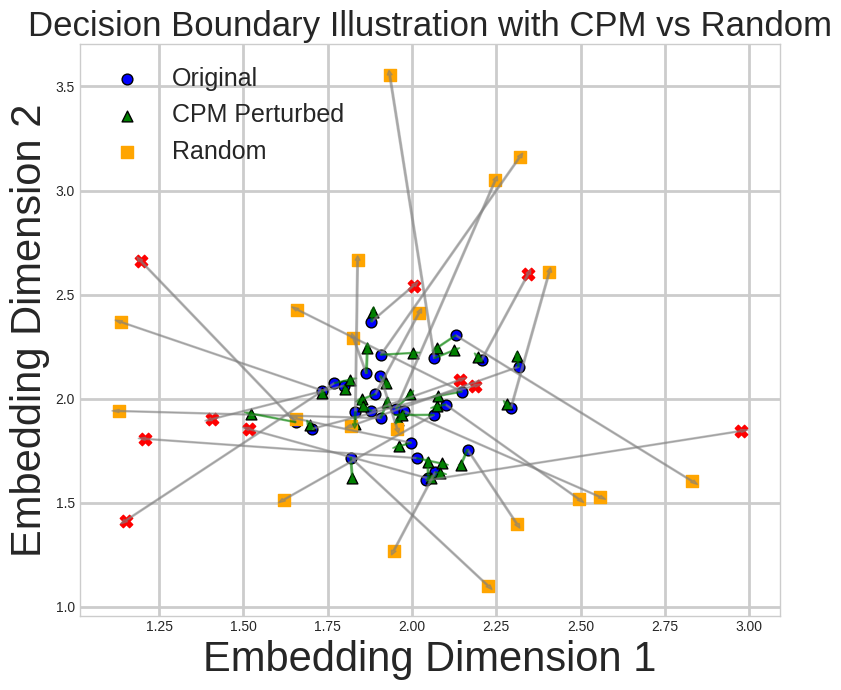}
        \caption{}
        \label{fig:decision_boundary}
    \end{subfigure}
    \hfill
    \begin{subfigure}{0.44\columnwidth}
        \centering
        \includegraphics[width=\columnwidth, height=4.2cm]{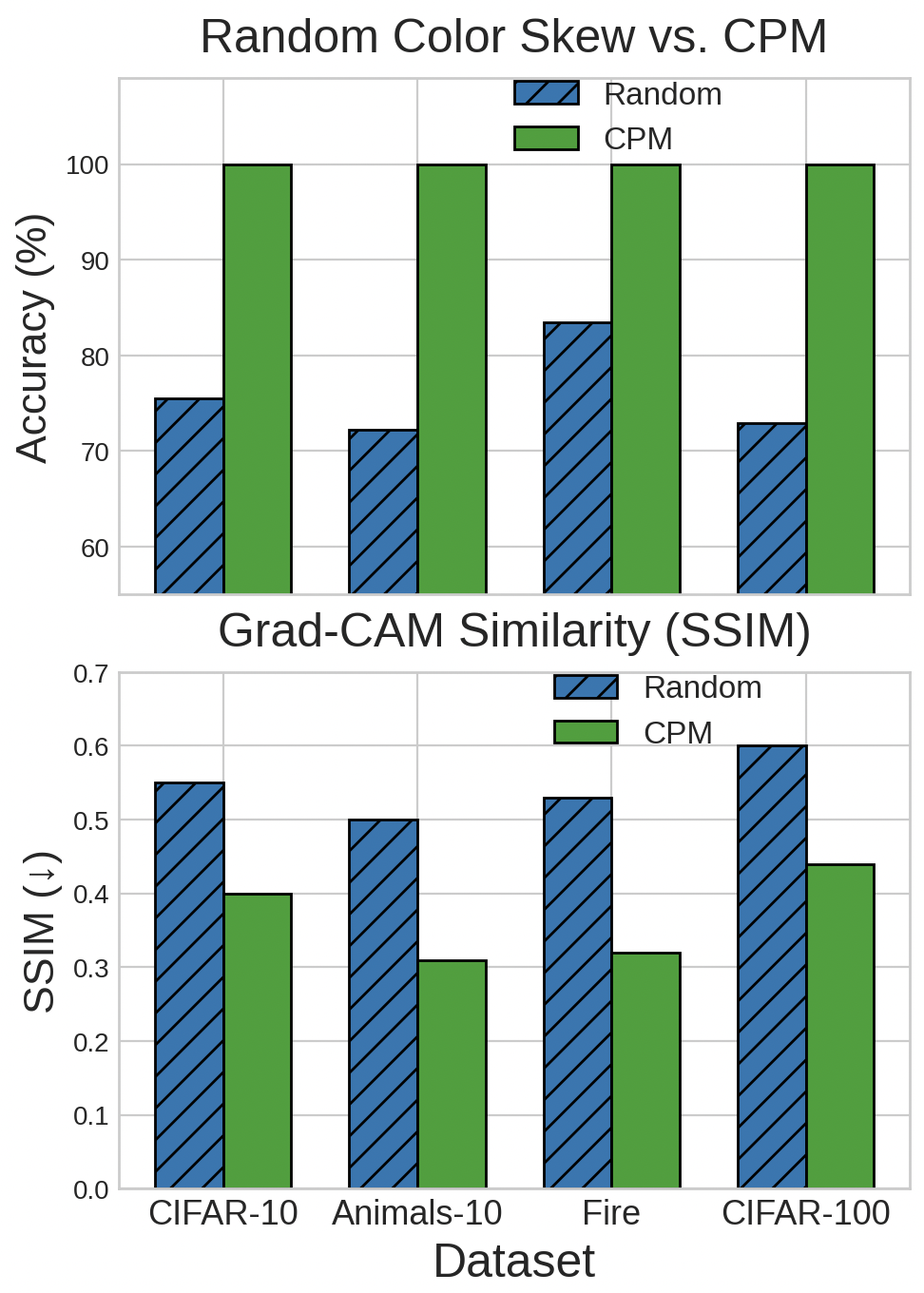}
        \caption{}
        \label{fig:bar_tradeoff}
    \end{subfigure}
    \caption{Comparative Analysis of CPM and Random Color skew. Results in (b) showing random color skew (averaged over random parameter selections) and CPM across datasets}
    \label{fig:cpm_vs_random_analysis}
\end{figure}

Figure~\ref{fig:cpm_vs_random_analysis} illustrates these effects. In the Fire dataset embedding (Subfigure~\ref{fig:decision_boundary}), CPM samples (green) remain in-class, while random ones often cross boundaries (orange/red). Subfigure~\ref{fig:bar_tradeoff} shows that CPM maintains 100\% accuracy while lowering SSIM, whereas random color skew that drives SSIM below 0.6 severely degrades accuracy. Thus, CPM preserves predictions while subtly altering explanations, unlike unstable random distortions.

\subsection{non-CAM based interpretability under CPM}
\vspace{-0.5 em}
While our primary focus is on Grad-CAM, we also evaluated CPM on non-CAM attribution methods, including Saliency~\cite{simonyan2013deep}, Integrated Gradients (IG)~\cite{sundararajan2017axiomatic}, and DeepLIFT~\cite{shrikumar2017learning} (Figure~\ref{fig:3methodCMP}). Results show that CPM consistently degrades these explanations as well, with low SSIM scores (e.g., 0.22 for Saliency, 0.51 for IG, 0.32 for DeepLIFT), despite predictions remaining unchanged. This confirms that the attack is not limited to CAM-based methods but extends broadly to gradient-based attribution techniques. The supplementary file provides extended comparisons across interpretability tools, showing that Grad-CAM is the most widely adopted and effective choice for human auditing.

\subsection{Ablation on Perturbation Operators}

To assess the contribution of each perturbation component in CPM, we perform an ablation study using all test samples from CIFAR-10. In table~\ref{tab:ablation_cpm}, the results show that individual operators are either too weak to cause saliency shift or too strong to flip prediction easily, highlighting the necessity of combining them in CPM. More details are available in the supplementary material.

\subsection{Proposed Attack Against Common FL Defenses}
\textbf{Robust Aggregation:}
Robust aggregation methods like Trimmed Mean, Median, and FLTrust detect poisoned updates via statistical anomalies. In contrast, CPM preserves accuracy and gradient magnitudes, making updates appear benign. As shown in Table~\ref{tab:robust_agg}, these defenses are only partially effective with CPM but still cause notable interpretability degradation. Though less severe than with plain FedAvg, SSIM scores remain low, and peak overlap drops. Achieving SSIM $< 0.5$ requires more rounds and higher adversary ratios, emphasizing the need for defenses that also monitor explanation fidelity.

\begin{table}[htbp]
\scriptsize
\setlength{\tabcolsep}{0pt}
\centering
\begin{tabular}{lccc}
\toprule
\textbf{Operator(s)} & \textbf{SSIM (↓)} & \textbf{Success (\%)} & \textbf{Failure Mode} \\
\midrule
Hue Shift           & 0.71 & 60 & Needs large shift → visible \\
Channel Rescaling   & 0.53 & 83 & Needs large shift → visible \\
Contrastive Jitter  & 0.71 & 52 & Breaks prediction at high jitter \\
CPM (combined)  & 0.478 & \textbf{100} & — \\
\bottomrule
\end{tabular}
\caption{Ablation of individual perturbation operators.}
\label{tab:ablation_cpm}
\end{table}

\begin{table}[htbp]
\centering
\small
\setlength{\tabcolsep}{2pt} 
\renewcommand{\arraystretch}{1}
\begin{tabular}{|p{3.3cm}|c|c|c|}
\hline
\textbf{Method} & \textbf{Acc. (\%)} & \textbf{SSIM (↓)} & \textbf{Peak Overlap} \\
\hline
FedAvg (baseline) \cite{mcmahan2017fedavg}    & 98.4 & 0.551 & 41.3 \\
Trimmed Mean \cite{wang2022federated}        & 97.9 & 0.667 & 58.2 \\
Median \cite{pillutla2019robust}              & 97.7 & 0.671 & 59.0 \\
FLTrust \cite{cao2020fltrust}             & 98.5 & 0.659 & 57.7 \\
\hline
\end{tabular}
\caption{CPM's effect persists across robust aggregation strategies.}
\label{tab:robust_agg}
\end{table}

\noindent\textbf{Benchmark defenses:}
While both adversarial training \cite{goodfellow2014explaining} and data augmentation \cite{shorten2019survey} aim to increase model robustness, they do not explicitly defend against interpretability attacks. As shown in Table~\ref{tab:defense_comparison}, these methods rely on altering predictions or labels to train against input perturbations, whereas CPM targets saliency directly while preserving model decisions. This unique property allows CPM to evade conventional defenses designed for prediction robustness, not explanation fidelity. Evaluation on each client shown in Table \ref{tab:baseline_fidelity} follows basic data augmentation before training. We exclude adversarial training in the training process as it relies on label-flipping attacks, while CPM preserves predictions and targets saliency, making such defenses ineffective.

\begin{figure}
    \centering
    \includegraphics[width=0.8\columnwidth, height=3.5cm]{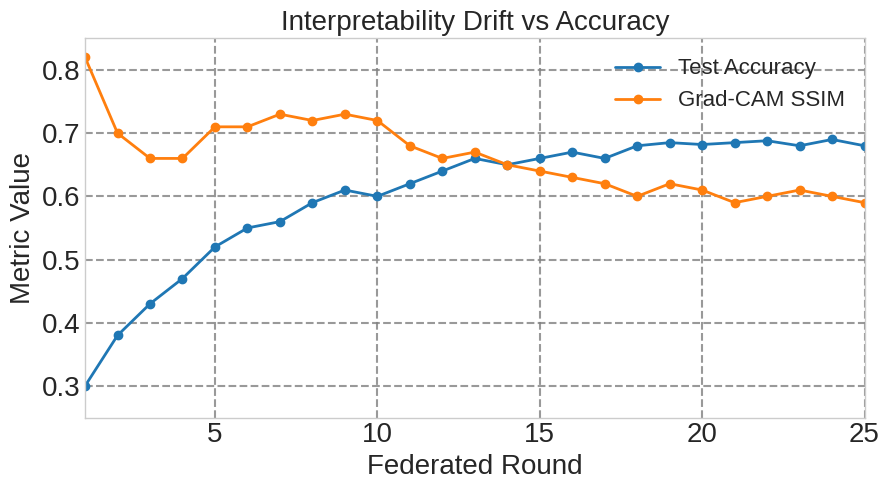}
    \caption{Interpretability Drift vs. Accuracy over FL rounds when attack samples are transferred across architectures.}
    \label{fig:CPM_limitation}
\end{figure}

\begin{table}[htbp]
\centering
\small
\renewcommand{\arraystretch}{1.2}
\setlength{\tabcolsep}{3pt} 
\begin{tabular}{|l|c|c|c|}
\hline
\textbf{Capability} & \textbf{Aug.} & \textbf{Adv. Train} & \textbf{CPM} \\
\hline
Preserves Accuracy              & \cmark & \cmark & \cmark \\
Targets Interpretability        & \xmark & \xmark & \cmark \\
Needs Label Manipulation        & \xmark & \cmark & \xmark \\
Optimized for Grad-CAM Dist.    & \xmark & \xmark & \cmark \\
\hline
\end{tabular}
\caption{Comparison of CPM with benchmark defenses.}
\label{tab:defense_comparison}
\end{table}


\subsection{Limitations and Future Work} 

While CPM is effective within the same model architecture, its transferability across architectures is limited. Attack samples generated on one model (e.g., MobileNet) show reduced effectiveness when applied to another (e.g., DenseNet121), with accuracy dropping despite Grad-CAM similarity decreasing (Figure~\ref{fig:CPM_limitation}). This highlights a limitation of CPM in cross-architecture settings and motivates future work on more transferable perturbations. 

Investigating defenses like differentiable saliency-aware adversarial training and data augmentation remains an important direction for future work. One possible defense is to monitor explanation consistency over rounds using SSIM or peak overlap on held-out samples. Another is to incorporate saliency alignment into the training objective, encouraging stability in attribution maps for inputs with unchanged predictions.

\section{Conclusion}

This work introduces a unique attack that degrades model interpretability in federated learning without affecting prediction accuracy. The proposed Chromatic Perturbation Module applies structured color-based perturbations to shift Grad-CAM explanations away from semantically meaningful regions while preserving model decisions. We demonstrate that CPM is effective across datasets, persists over multiple rounds of aggregation in FL, and evades standard defenses, including data augmentation and robust aggregation. These results expose an overlooked vulnerability that interpretability itself can be an attack surface. Future directions include developing defenses that explicitly preserve explanation fidelity.

\section*{Acknowledgment}
This material is based upon work supported by the U.S. Department of Energy, Office of Science, Office of Advanced Scientific Computing Research under Contract No. DE-AC05-00OR22725. This manuscript has been co-authored by UT-Battelle, LLC under Contract No. DE-AC05-00OR22725 with the U.S. Department of Energy. The United States Government retains and the publisher, by accepting the article for publication, acknowledges that the United States Government retains a non-exclusive, paid-up, irrevocable, world-wide license to publish or reproduce the published form of this manuscript, or allow others to do so, for United States Government purposes. The Department of Energy will provide public access to these results of federally sponsored research in accordance with the DOE Public Access Plan (http://energy.gov/downloads/doe-public-access-plan). Additionally, this work was also supported by the US National Science Foundation (NSF) under grant CNS-2038922.

\bibliographystyle{IEEEtran}
\bibliography{ref}

\end{document}